\documentclass{article}

\usepackage{microtype}
\usepackage{graphicx}
\usepackage{subfig}
\usepackage{booktabs} 

\usepackage{hyperref}

\usepackage[accepted]{icml_fmwild}

\usepackage{amsmath}
\usepackage{amssymb}
\usepackage{mathtools}
\usepackage{amsthm}
\usepackage{aas_macros} 

\usepackage[capitalize,noabbrev]{cleveref}

\theoremstyle{plain}

\theoremstyle{definition}

\theoremstyle{remark}

\usepackage[textsize=tiny]{todonotes}

\newcommand{\teff}{\ensuremath{T_\mathrm{eff}}}
\newcommand{\logg}{\ensuremath{\log g}}
\newcommand{\xh}[1]{\ensuremath{[\mathrm{#1/H}]}}

\newcommand{\gaia}{\emph{Gaia}}
\newcommand{\bp}{\ensuremath{G_\mathrm{BP}}}
\newcommand{\rp}{\ensuremath{G_\mathrm{RP}}}
\newcommand{\bprp}{\ensuremath{G_\mathrm{BP}-G_\mathrm{RP}}}

\definecolor{darkgreen}{rgb}{0.0, 0.7, 0.0}

\definecolor{darkblue}{rgb}{0.0, 0., 0.7}

\icmltitlerunning{Estimating Probability Densities with Transformer and Denoising Diffusion}

\begin{document}

\twocolumn[
\icmltitle{Estimating Probability Densities of Tabular Data using a Transformer Model combined with Denoising Diffusion}



\icmlsetsymbol{equal}{*}

\begin{icmlauthorlist}
\icmlauthor{Henry W. Leung}{astro,dunlap}
\icmlauthor{Jo Bovy}{astro,dunlap}
\icmlauthor{Joshua S. Speagle}{astro,dunlap}
\end{icmlauthorlist}

\icmlaffiliation{astro}{David A. Dunlap Department of Astronomy and Astrophysics, University of Toronto, 50 St. George Street, Toronto, Ontario, M5S 3H4, Canada}
\icmlaffiliation{dunlap}{Dunlap Institute for Astronomy and Astrophysics, University of Toronto, 50 St. George Street, Toronto, Ontario, M5S 3H4, Canada}

\icmlcorrespondingauthor{Henry W. Leung}{henrysky.leung@utoronto.ca}
\icmlcorrespondingauthor{Jo Bovy}{bovy@utoronto.ca}
\icmlcorrespondingauthor{Joshua S. Speagle}{j.speagle@utoronto.ca}

\icmlkeywords{Machine Learning, ICML}

\vskip 0.3in
]



\printAffiliationsAndNotice{}  

\begin{abstract}
    Transformers are often the go-to architecture to build foundation models that ingest a large amount of training data. But these models do not estimate the probability density distribution when trained on regression problems, yet obtaining full probabilistic outputs is crucial to many fields of science, where the probability distribution of the answer can be non-Gaussian and multimodal. In this work, we demonstrate that training a probabilistic model using a denoising diffusion head on top of the Transformer provides reasonable probability density estimation even for high-dimensional inputs. The combined Transformer+Denoising Diffusion model allows conditioning the output probability density on arbitrary combinations of inputs and it is thus a highly flexible density function emulator of all possible input/output combinations. We illustrate our Transformer+Denoising Diffusion model by training it on a large dataset of astronomical observations and measured labels of stars within our Galaxy and we apply it to a variety of inference tasks to show that the model can infer labels accurately with reasonable distributions.
\end{abstract}

\section{Introduction}\label{sec:ddpm_intro}

Many scientific fields are becoming increasingly data-driven and use machine learning for a large variety of analysis and synthesis tasks. For example, in the field of astronomy, many areas benefit from large-scale surveys across multiple areas such as spectroscopy, photometry, and time-domain observations. Good examples of this are the \gaia\ survey, which mostly produces tabular data, the Sloan Digital Sky Survey (SDSS;  \citealt{2017AJ....154...28B,2017arXiv171103234K}) spectroscopic surveys and The Dark Energy Camera Legacy Survey (DECaLS; \citealt{2019AJ....157..168D}) imaging surveys, and the upcoming Vera C. Rubin Observatory (LSST; \citealt{2019ApJ...873..111I}) which will produce multi-band images and time-series data for billions of objects. In order to build a machine learning model for astronomy that is useful across a wide range of use cases, such a model needs to be trained at scale across all of these datasets.

Foundation models are an active area of research in a wide range of areas, but for our purposes applications in scientific areas are most interesting. There they are showing great promise for revolutionizing science as they are able to leverage commonalities between different scientific areas (e.g., \citealt{2023arXiv231002994M}) and because they obey similar scaling laws as found in natural language processing, allowing these models to outperform traditional methods when trained with large amounts of data \citep{2024arXiv240402973W}.

Transformers are the currently most powerful and flexible architecture for building foundation models, especially in natural language because the attention mechanism used in Transformers can capture contextual dependencies within a sequence of data and can therefore predict the missing/next words. But in the astronomical data context where the learning task is regression, they only predict scalars for each feature with perhaps a scalar predictive uncertainty to represent the width of a Gaussian uncertainty distribution. While the Gaussian assumption is often a good assumption in science due to the central limit theorem, it can be untrue/uninformative especially when using foundation models performing a wide range of tasks on labels whose distributions can be non-Gaussian and multi-modal.

To infer probability densities for the output instead of scalars, we need to incorporate generative models on top of the Transformers, such that the model can learn the underlying distribution of the training data abd can generate plausible samples. In our model, we want the model to generate plausible samples based on conditions from Transformers. Possible choice of generative models for our application include Normalizing Flows (NF; \citealt{2015arXiv150505770J}) or a Denoising Diffusion Probabilistic Model (DDPM; \citealt{2020arXiv200611239H}). Here we chose DDPM because of its flexibility of inferring a wide range of distributions with a single model. DDPM takes samples from the training data distribution, incrementally adding noise in order to corrupt the data, and the role of DDPM is to gradually de-noise the data. A population of new samples can be generated from de-noising random noise (often normal) with the trained DDPM. 

The usage of Transformer and DDPM together is not a new idea and has been applied in various ways and various tasks. For example, in natural language settings like \citet{2024arXiv240520519K} for diagram generation using a Transformer with DDPM head, or using the Transformer architecture within DDPM like \citet{2024arXiv240409636G} for simulation-based inference and \citet{2022arXiv221209748P} for natural language image generation.

\begin{figure}
  \centering
  \includegraphics[width=0.8 \linewidth]{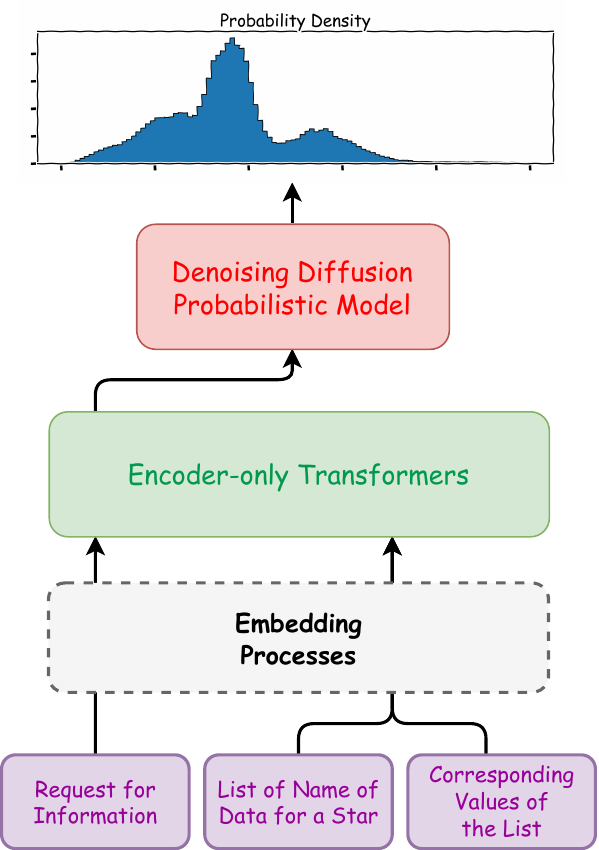}
  \caption[High level model architecture of our Transformer+Denoising Diffusion model]{High level model architecture of our Transformer+Denoising Diffusion foundation model. The goal of the model is to estimate the probability density of an output based on a list of input scientific data and what unknown data is being requested. The role of the Denoising Diffusion Probabilistic Model (DDPM) is to turn the hidden state at the first position from the Transformer to a probability density distribution.}
  \label{fig:ddpm_archetiture}
\end{figure}

In this work, we extend the work of \citet{2024MNRAS.527.1494L} (thereafter LB24) on a scientific foundation model by incorporating a DDPM head on top of a Transformer trained in a regression setting (as opposed to logistic regression used in natural language), trained with a small set of astronomical data of stars in our Galaxy. We demonstrate that the model can provide reasonable probability density estimation when given arbitrary combinations of inputs and outputs, hence enhancing the information output. Our technique should be immediately applicable to other foundation models especially in science and with model with higher dimensional output.
\begin{figure}
  \centering
  \includegraphics[width=0.9 \linewidth]{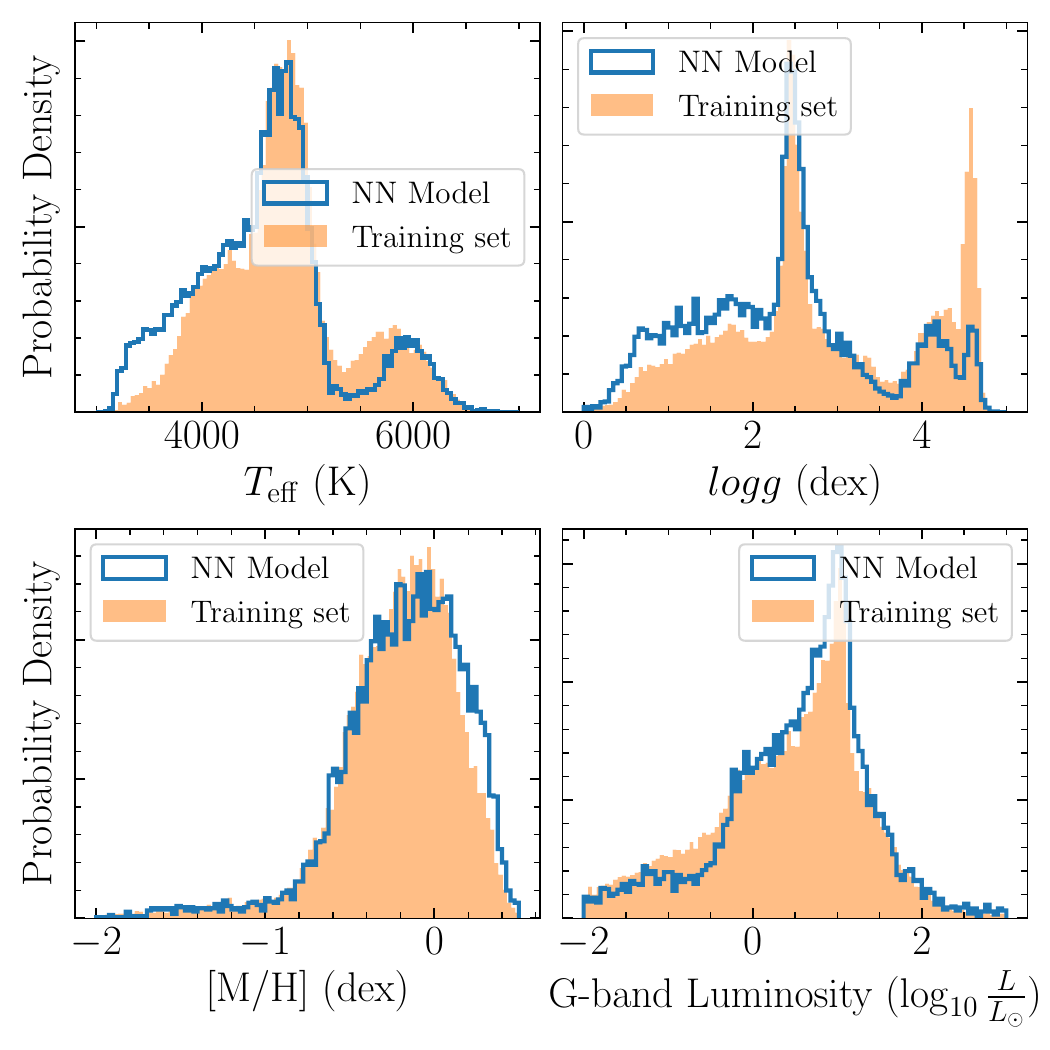}
  \caption{Probability density of different surface temperature \teff\ (leftmost), surface gravity \logg\ (middle left), metallicity \xh{M} (middle right) and G-band luminosity (rightmost) of stars based on no conditions. The blue solid lines show the probability density from our model while the orange colored histogram shows the probability density from the training set. This figure demonstrates that our model learns the training set distribution on various labels when there is no condition.}
  \label{fig:ddpm_no_conditions}
\end{figure}

\section{Motivation}\label{sec:ddpm_motivation}

Neural network models trained on discriminative tasks in general do not provide probability density estimation for the outputs, because the learning objective is simply minimizing the difference between the model prediction and the ground truth. There is no mechanism in the model to estimate the probability distribution.

As demonstrated in LB24, it is possible to train a model where one can give observations and known properties of a star, and then request a quantity that you do not know about that star. For example, one can give a spectrum of a star and ask the model for the iron abundance, which is an easy, unambiguous task in astronomy. But when the model is asked to do an ambiguous task (i.e. there are multiple ground truths) like inferring the surface gravity \logg\ from surface temperature \teff\ (one can refer to the right panel of \cref{fig:ddpm_kiel_generated} for the distribution of stars in this space), the model predicts a mean surface gravity with large uncertainty. While this is correct when assuming Gaussian uncertainty, because there can be a wide range of possible surface gravities, the output is uninformative since the true probability distribution of surface gravity is composed of two widely-separated narrow peaks. This is why it is crucial to have a model that can output non-trivial probabilities while keeping the flexibility of the Transformer.

\begin{figure}
  \centering
  \includegraphics[width=0.99 \linewidth]{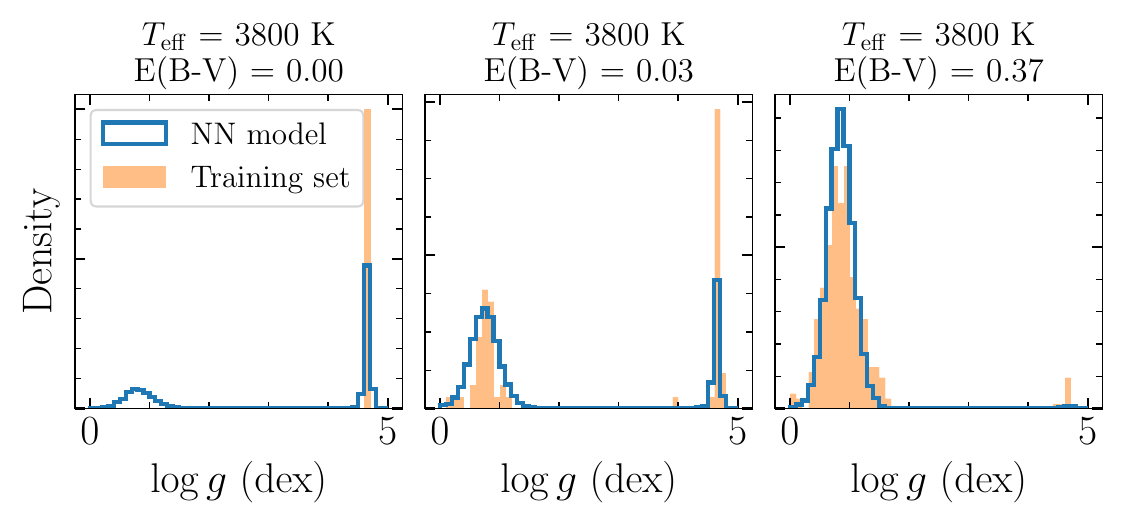}
  \caption{Probability density of surface gravity \logg\ given different surface temperature \teff\ and reddening $E(B-V)$ (hence extinction) of stars. All panels show the probability density from the model (solid blue lines) and the training set (orange filled area) based on the condition provided in the panel's title. As reddening increases, a higher proportion of stars are expected to be intrinsically bright giant stars rather than intrinsically dim dwarf stars. This figure demonstrates that our model learns the correct output distribution of various labels when there are only a few conditions leading to an ambiguous answer not captured by traditional Transformer-only models.}
  \label{fig:ddpm_teff_extinction}
\end{figure}

\section{Model Implementation}\label{sec:ddpm_model}

We adopt a similar architecture, embedding process, and training procedure as in LB24, but modify their approach in a few aspects. Instead of an encoder-decoder Transformer, here we use an encoder-only architecture similar to Google's BERT \citep{2018arXiv181004805D} to reduce of the number of attention mechanisms needed to be performed to reduce the computational cost. Moreover to alleviate the issue of having a limited context size during inference, the encoder-only Transformer architecture used here allows the context size during inference to be much longer than the one we have trained on, because we do not require positional-encoding here. The model accepts a list of data points that consist of the name of the data as well as their magnitude (e.g., [\teff,$7000\,\mathrm{K}$]); the first position in the input sequence is always the name of the data being requested without the magnitude.

We use the same ``non-linear embedding'' process as described in LB24, which embeds data of kind $x$ through
\begin{equation}\label{eq:embedding}
    y_x = f(w_x \cdot M_x) + w_{b, x}
\end{equation}
where $y_x$ is the final vectorized data for a particular kind of data $x$ that will be fed into the encoder-only Transformer. The function $f$ is a typical activation function used in neural networks, $w_x$ are the embedding weights, $M_x$ is the magnitude of the data, and $w_{b, x}$ is a bias weight; all of these are particular to the kind of data $x$.

As shown in \cref{fig:ddpm_archetiture}, the outputs from the Transformer passed to the DDPM head are from the hidden state at the position of the request token (similar to \texttt{[CLS]} in Google's BERT classification token); the rest of the hidden states are not used. The DDPM head de-noises samples from a unit Gaussian distribution where the denoising process is controlled by the hidden state at the position of the request token. During inference, the probability distribution is generated by denoising samples from an unit normal distribution based on the output from the Transformer.

\begin{figure}
  \centering
  \includegraphics[width=\linewidth]{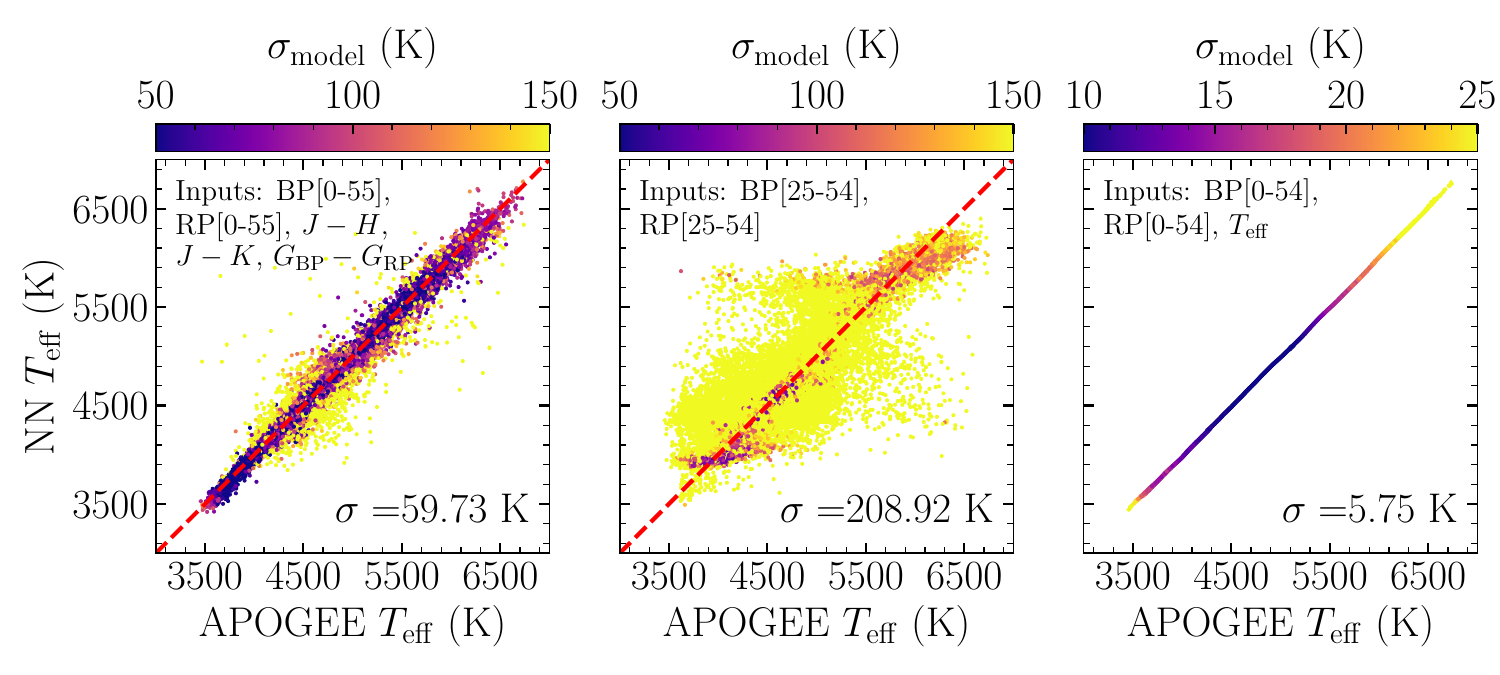}
  \caption{Inference of surface temperature \teff\ from different combinations of input data from the testing set. NN \teff\ is the median of the output probability distribution with the robust standard deviation (i.e., $1.4826 \times$median absolute deviation) of the distribution as the output uncertainty represented by colors. In the left panel, the input contains the whole Gaia XP spectra with colors (113 data point in total which is almost double the 64 context size used during training) and the \teff\ is accurate with reasonable uncertainty. In the middle panel, the input contains the most uninformative part of the Gaia XP spectra and the model prediction of \teff\ also suffers with large uncertainty as expected. In the right panel, the same input as the middle panel is given, but in addition \teff\ is mixed into the inputs, in this case the model predicts almost perfect \teff\ with very low uncertainty.}
  \label{fig:ddpm_teff_accurancy}
\end{figure}

\section{Datasets and Training}\label{sec:ddpm_data_train}

To illustrate the model, we use the same training and testing data set as in LB24. The data sets are small enough for quick proof-of-concept purposes but big enough for a meaningful foundation model to be fit, and the result can be directly compared to their work. The data set consists of Gaia DR3 photometry and XP spectra \citep{2023A&A...674A...2D, 2023A&A...674A...3M}, 2MASS photometry \citep{2006AJ....131.1163S}, and stellar properties derived from high-resolution, high signal-to-noise-ratio near-infrared APOGEE DR17 \citep{2022ApJS..259...35A} spectra for $\approx 100,000$ stars. The majority of these stars are sub-giants and red giants. Further details of these data are given in \cref{sec:ddpm_astro_data}.

In the data set, each star has a row of available observations in which missing data (represented by \texttt{NaN}) are replaced by a special padding token (the same idea used to mask empty spaces in sentences), which is masked automatically in the attention layer. Similar to LB24, we pick a random set of data for each star as input and a random one as output, which may or may not be included in the input already even for stars in the same batch during training. Hyper-parameters used in training are described in \cref{sec:ddpm_hyperparameters}.

\begin{figure}
  \centering
  \includegraphics[width=0.8 \linewidth]{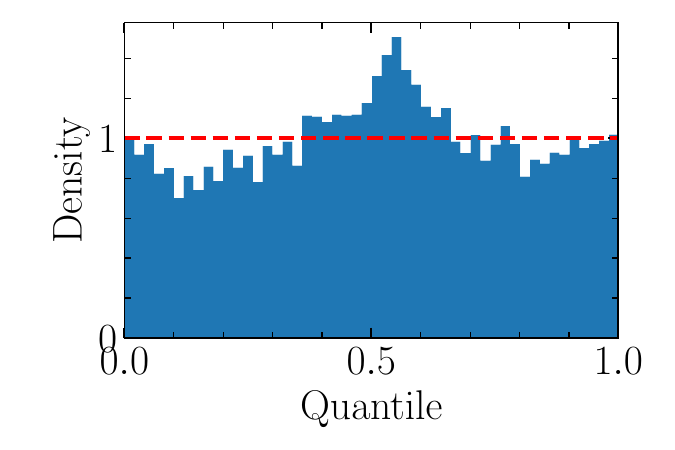}
  \caption{The distribution of the quantile a which the ground truth is found in the probability density distribution inferred from our model. Similar to the training procedure, we randomly select a subset of data as input and a random label as output for each star. The blue colored histogram show the quantile distribution while the red dotted line represents the uniform distribution, which the quantiles should follow if the model probability distributions were exactly correct. The quantile distribution closely follows the red dotted line with a slight over-concentration around the $50\%$ quantile, indicating that we are slightly overestimating the uncertainty in the outputs.}
  \label{fig:ddpm_quantile_test}
\end{figure}

\section{Training Hyper-parameters}\label{sec:ddpm_hyperparameters}

The model used for training consists of total 3.7 million trainable parameters which is slightly less than half of the 8.8 million parameters used in LB2024. This is also many order of magnitudes less than those used in commercial models which have billions of parameters because this is a proof-of-concept. There are 118 of type of data and each embedding we have used is 192-dimensional. The encoder-only Transformer consists of 2 multi-head attention layers where each layers has 24 attention heads such that each head corresponds to 8-dimension in the embedding.

For the DDPM head, there are 1.8 million trainable parameters which is about half of the total 3.7 million parameters for the whole model. There are 120 diffusion steps, that is we generate a new sample of data by denoising 120 times from pure Gaussian noise. A sigmoid noise schedule $\beta$ is used, which is defined as follow

\begin{equation}\label{eq:sigmoid_beta}
\beta_t = \beta_{\text{start}} + (\beta_{\text{end}} - \beta_{\text{start}}) \cdot \frac{1}{1 + e^{-t}}
\end{equation}

where $\beta_\text{start}$ and $\beta_{\text{end}}$ are $0.00001$ and $0.1$, respectively and $t$ are uniformly sampled from $-6$ to $6$. The \texttt{AdamW} optimizer is used to train the whole model with an initial learning rate of $0.001$ with a \texttt{CosineAnnealingWarmRestarts} learning rate scheduler with final learning rate of $1\times10^{-10}$; each cycle spans 1024 epochs for 10240 epochs in total during training. The training process took place with a single NVIDIA A100SXM4 for about 20 hours.
\section{Results}\label{sec:ddpm_result}

To validate our model, we perform variety of tests in different scenarios. First, we test the model's ability to infer probability density distributions when no condition is provided, which essentially checks if the model can recover the training set distribution (\cref{fig:ddpm_no_conditions}). The second test is to see if the model can recover probability density distributions when there are conditions (\cref{fig:ddpm_teff_extinction}). The third test is to check if the median and standard deviation of the distribution are accurate and provide reasonable uncertainty estimation (\cref{fig:ddpm_teff_accurancy}). The final test checks how good the output probability densities are, by checking that the ground truth is found at uniformly distributed quantiles of the model probability densities (\cref{fig:ddpm_quantile_test}).

Overall, our model works reasonably well when we use the probability density distribution to estimate population of stars (see \cref{fig:ddpm_no_conditions} and \ref{fig:ddpm_teff_extinction}). Our model also works reasonably well when one simply uses the mean as the final prediction and the standard derivation as the output uncertainty as shown in \cref{fig:ddpm_teff_accurancy}, similar to Figure 4 in LB24. The prediction is accurate and the estimated uncertainties are reasonable, with predictions far from the ground truth also being those with large uncertainty. The model also perform inference on arbitrary combinations of input and outputs;  \cref{fig:ddpm_quantile_test} shows the quantile of the ground truth within the model predicted probability distribution. The uniformity in this quantile distribution demonstrates that the model provides a reasonable uncertainty in general.

\begin{figure}
  \centering
  \includegraphics[width= \linewidth]{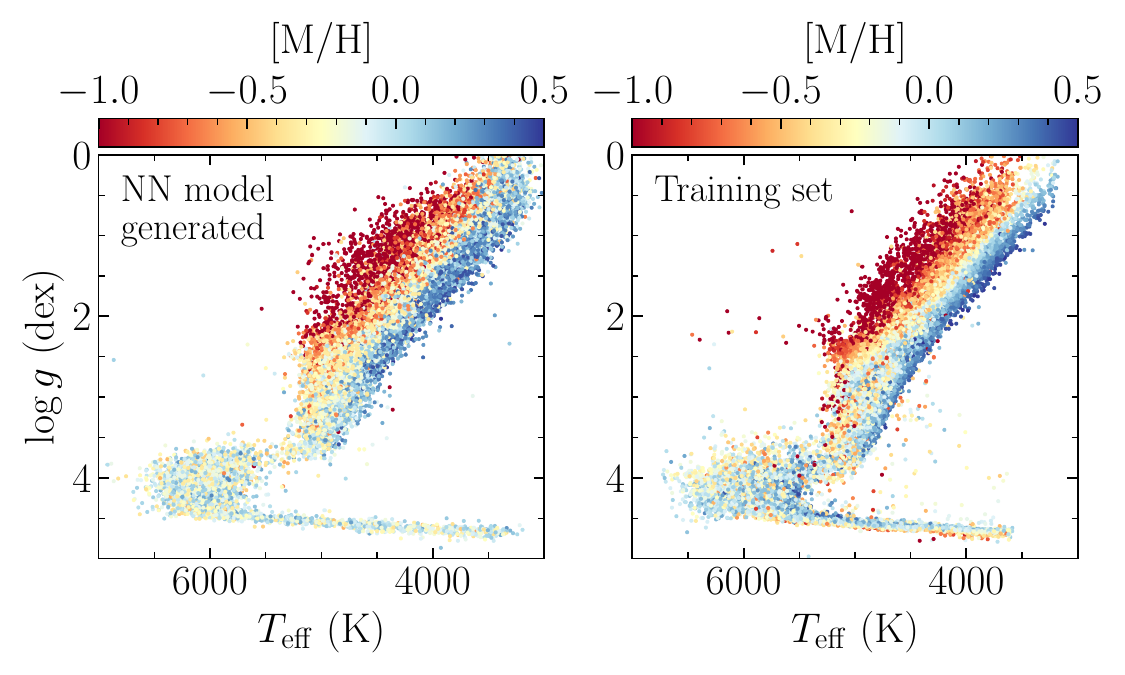}
  \caption{Surface temperature \teff\ and surface gravity \logg\ diagram (a.k.a. Kiel Diagram) reconstructed from with our model on the left with the ground-truth distribution on the right. Because our model only output one-dimensional distributions, the diagram was generated by first requesting the distribution of metallicity, then ask for the distribution of \teff\ for each metallicity sample and finally ask for the distribution of \logg\ for each pair of \teff\ and metallicity samples. Our model's reconstruction represents the training set distribution well.}
  \label{fig:ddpm_kiel_generated}
\end{figure}

Even though the model presented here generates only one dimensional distributions, the model is capable of providing multi-dimensional distributions by using the flexibility of the Transformer to generate a single dimension at a time sequentially. An example is shown in \cref{fig:ddpm_kiel_generated} where the model distribution in the three-dimensional space of surface temperature$-$surface gravity$-$metallicity. We first ask the model to generate the metallicity \xh{M} distribution, then use the \xh{M} samples to generate surface gravity samples, and use pairs of samples to generate surface temperature samples.  This procedure allows one to generate probability distributions for \emph{any} combination of output quantities, thus solving the combinatoric problem of learning large numbers of high-dimensional probability distributions.

\section{Conclusion}\label{sec:ddpm_conclusion}

In this work, we have trained an encoder-only Transformer with a denoising diffusion probabilistic model head on top with a small set of astronomical dataset of stars in our Galaxy. We demonstrated that the model can infer probability densities for the outputs reasonably well when the model is asked to recover the training set probability density distribution of various labels or when asking the model for conditional probability density distributions, making this model a very flexible density function emulator. The prediction and uncertainty of the distributions from the model are accurate and reasonable, making the model useful when the user only wants a scalar prediction and a scalar uncertainty instead of a whole distribution. The approach we demonstrate here can be used in other scientific foundation models in regression settings to make the model's outputs more informative.
\newpage
\section*{Software and Data}

Python is used in this work and all packages used here are open-source software include \texttt{pytorch}, \texttt{numpy}, \texttt{matplotlib}, \texttt{pandas}, \texttt{h5py}, \texttt{tqdm}, \texttt{astropy}, \texttt{mwdust}, \texttt{GaiaXPy}, \texttt{gaiadr3\_zeropoint}, \texttt{MyGaiaDB} \& \texttt{mwdust}. All astronomical data used in this work are public from the Sloan Digital Sky Survey and European Space Agency Gaia mission.

The code for this work is available at \url{https://github.com/henrysky/stars_foundation_diffusion}.

\section*{Acknowledgements}

HL and JB acknowledge financial support from the Natural Sciences and Engineering Research Council of Canada (NSERC; funding reference number RGPIN-2020-04712). 

This research was enabled in part by support provided by Compute Ontario (\url{https://www.computeontario.ca/}) and the Digital Research Alliance of Canada (\url{alliancecan.ca}).

This project is supported by the Data Sciences Institute at the University of Toronto.

Funding for the Sloan Digital Sky Survey IV has been provided by the Alfred P. Sloan Foundation, the U.S. Department of Energy Office of Science, and the Participating Institutions. SDSS-IV acknowledges support and resources from the Center for High Performance Computing at the University of Utah. The SDSS website is www.sdss.org.

This work has made use of data from the European Space Agency (ESA) mission
{\it Gaia} (\url{https://www.cosmos.esa.int/gaia}), processed by the {\it Gaia}
Data Processing and Analysis Consortium (DPAC,
\url{https://www.cosmos.esa.int/web/gaia/dpac/consortium}). Funding for the DPAC
has been provided by national institutions, in particular the institutions
participating in the {\it Gaia} Multilateral Agreement.

\section*{Impact Statement}

The work presented here can be applied to many other field of science.
This paper presents work whose goal is to advance Foundation model in scientific fields. There are minimal societal consequences and ethical concerns of our work.

\bibliography{main}
\bibliographystyle{icml2024}

\newpage
\appendix
\onecolumn

\section{Demonstration with California Housing Data}\label{sec:ddpm_california}

To demonstrate that our method works in a simple setting with a standard dataset, we have trained our model on the much simpler California Housing dataset \citep{KelleyPace1997} with only 20,640 data points and 9 features included in \texttt{scikit-learn}. We treat this dataset as tabular data and follow the same training procedure as in the main text (i.e., randomly select a subset of features as input and one feature as output during training for each house in each batch). The model consists of 47,425 trainable parameters including the 16-dimensional embeddings where 21,377 of those parameters comes from the DDPM head.

\begin{figure*}
  \centering
  \includegraphics[width=0.85\linewidth]{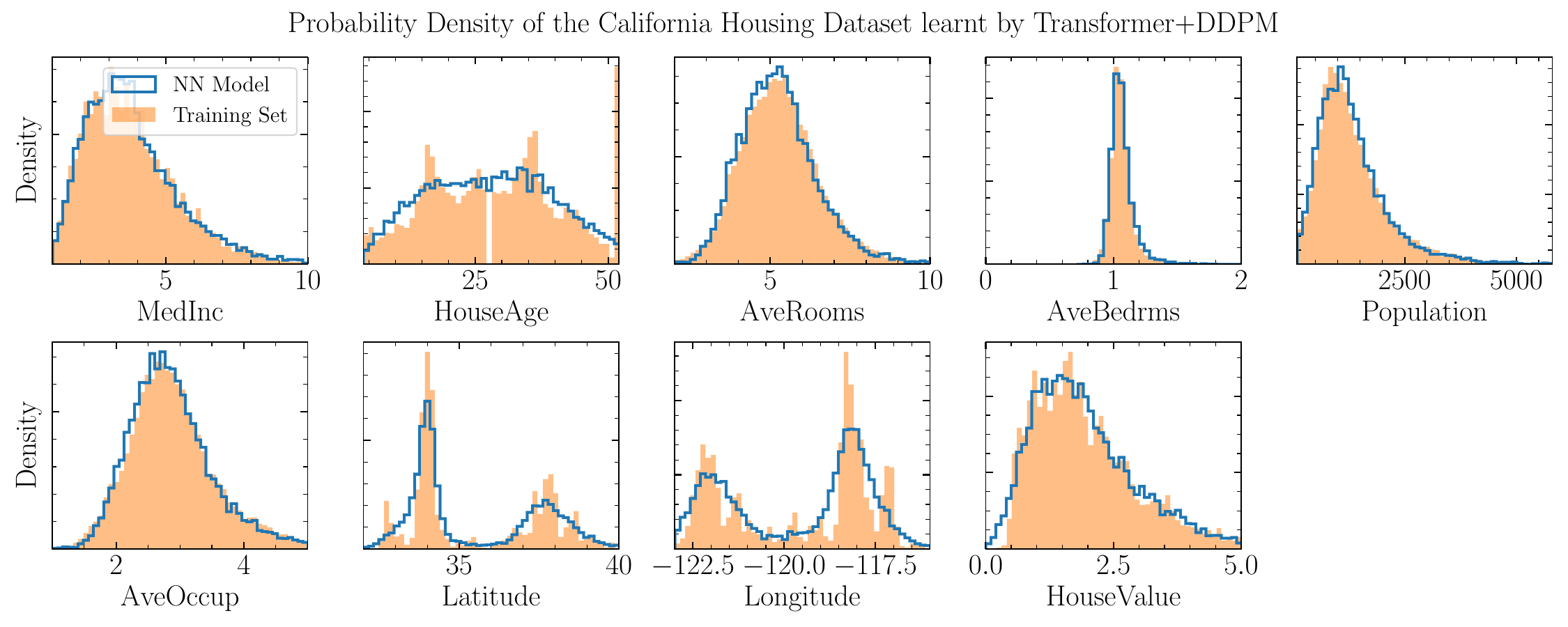}
  \caption{Probability density of the dataset based on no conditions (i.e., request values based only on padding). The orange filled areas are the density of the training set while the blue lines show the model prediction. In general, the predictions closely follow the training set distributions.}
  \label{fig:ddpm_california_training}
\end{figure*}

\begin{figure*}
  \centering
  \includegraphics[width=0.35\linewidth]{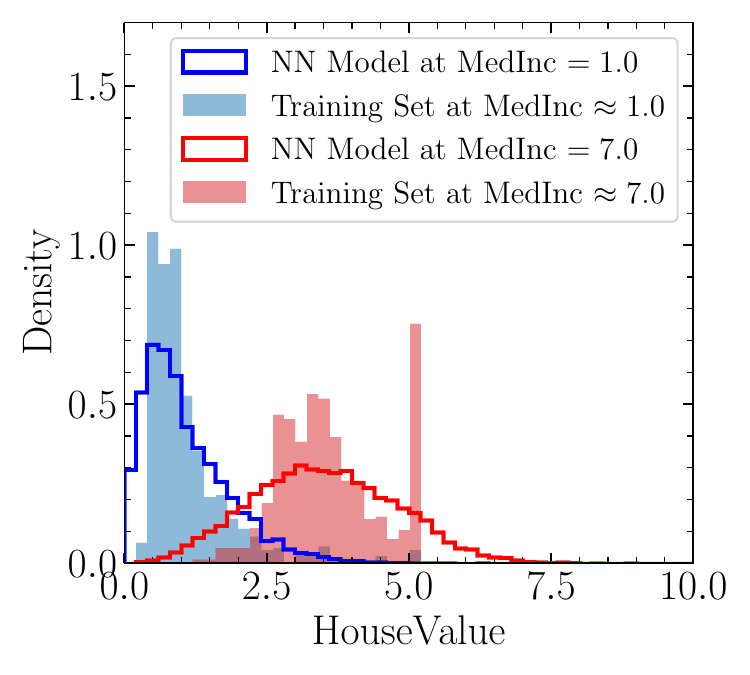}
  \caption{Conditional probability density of house value when providing the owner's median income. The colored areas are the density of the training set while the solid colored lines show the model prediction. People with higher median income tend to own a more expensive house as expected.}
  \label{fig:ddpm_california_conditional}
\end{figure*}

\cref{fig:ddpm_california_training} demonstrates that the model successfully predicts the distribution of all 9 features in the training set and \cref{fig:ddpm_california_conditional} shows that the model also can provide conditional probability distributions. In both cases, the distributions look reasonable and follow the training set.

\section{Astronomical Datasets}\label{sec:ddpm_astro_data}

Gaia XP spectra are low-resolution optical to near-infrared ($R\sim30$ to $100$, $330$ to $1050 \,\mathrm{nm}$; \citealt{2021A&A...652A..86C}) spectra obtained from the Blue and Red Photometers (BP/RP) aboard the Gaia spacecraft. Unlike usual higher resolution stellar spectra, the Gaia XP spectra were released as 110 coefficients of an orthogonal basis function expansion, where lower-order coefficients explain large-scale features of the spectra (hence more information like \teff) and higher-order coefficients explain small-scale features including the noise. We simply treat each coefficient as a kind of data. We normalize the XP coefficients by the Gaia $G$-band apparent flux.

Stellar parameter labels like the surface temperature \teff, surface gravity \logg, and the overall metallicity \xh{M} are from the SDSS APOGEE data release 17 \citep{2017AJ....154...28B, 2022ApJS..259...35A}. APOGEE is a high-resolution ($R \sim 22,000$), high signal-to-noise ($\>100$ per pixel typically) spectroscopic survey in both northern and southern hemisphere in the near infrared H-band wavelength region of $1.5-1.7 \mu m$. The stellar parameter labels are derived from APOGEE spectra using the APOGEE Stellar Parameter and Chemical Abundances Pipeline \citep{2016AJ....151..144G}.

Photometry $J$, $H$, $K$ band in infrared are from Two Micron All Sky Survey \citep{2006AJ....131.1163S} and \bp, \rp\ from Gaia spacecraft, $J-H$, $J-K$ and \bprp\ are simply the difference between these photometric bands. Reddening $E(B-V)$ are calculated with a three-dimensional extinction map \texttt{Combined19} \citep{2016ApJ...818..130B} based on three-dimensional position (i.e. two-dimensional coordinates on the sky as well as distance) of stars relative to Earth.

\end{document}